\documentclass{article}
\usepackage[preprint]{neurips_2023}
\usepackage[utf8]{inputenc} 
\usepackage[T1]{fontenc}    
\usepackage{hyperref}       
\usepackage{url}            
\usepackage{booktabs}       
\usepackage{amsfonts}       
\usepackage{nicefrac}       
\usepackage{microtype}      
\usepackage{xcolor}         
\usepackage{graphicx}
\setcitestyle{square, authoryear}
\newcommand\secref[1]{Section~\ref{#1}}
\newcommand\figref[1]{Figure~\ref{#1}}

\title{The Case Against Explainability}
\author{%
  Hofit Wasserman Rozen  \\
  Faculty of Law\\
  Tel-Aviv University\\
  Israel\\
  \And
  Niva Elkin-Koren \\
  Faculty of Law\\
  Tel-Aviv University\\
  Israel\\
  \And
  Ran Gilad-Bachrach \\
  Faculty of Engineering\\
  Tel-Aviv University\\
  Israel
}
\begin{document}
\maketitle
\begin{abstract}

As artificial intelligence (AI) becomes more prevalent there is a growing demand from regulators to accompany decisions made by such systems with explanations. However, a persistent gap exists between the need to execute a meaningful right to explanation vs. the ability of Machine Learning systems to deliver on such a legal requirement. The regulatory appeal towards "a right to explanation" of AI systems can be attributed to the significant role of explanations, part of the notion called reason-giving, in law. Therefore, in this work we examine reason-giving's purposes in law to analyze whether reasons provided by end-user Explainability can adequately fulfill them. 
We find that reason-giving's legal purposes include: (a)~making a better and more just decision, (b)~facilitating due-process, (c)~authenticating human agency, and (d)~enhancing the decision makers' authority. Using this methodology, we demonstrate end-user Explainabilty's inadequacy to fulfil reason-giving's role in law, given reason-giving's functions rely on its impact over a human decision maker. Thus, end-user Explainability fails, or is unsuitable, to fulfil the first, second and third legal function. In contrast we find that end-user Explainability excels in the fourth function, a quality which raises serious risks considering recent end-user Explainability research trends, Large Language Models' capabilities, and the ability to manipulate end-users by both humans and machines. Hence, we suggest that in some cases the right to explanation of AI systems could bring more harm than good to end users. Accordingly, this study carries some important policy ramifications, as it calls upon regulators and Machine Learning practitioners to reconsider the widespread pursuit of end-user Explainability and a right to explanation of AI systems. 
\end{abstract}
\section{Introduction}
As AI systems increasingly take up the role of assisting, and at times replacing, human decision makers \citep{Kaminski2021A}, there is a growing public call for establishing a right to receive an explanation to outcomes generated by automated decision-making processes. Purportedly originating from the General Data Protection Regulation (GDPR) \citep{GDPR} and known as the "legal right to explanation"\citep{Goodman2017}, this right is often portrayed as one tool in the regulatory toolkit for creating, deploying, and monitoring ethical and accountable AI systems, mitigating the potential breach of fundamental principles of the rule of law, such as transparency and accountability \citep{HILDEBRANDT} and protecting human rights. Simultaneously, and in correlation with the introduction of more complex AI, Explainability has triggered a growing interest within the Machine Learning (ML) community. Tasked with providing explanations to complex predictions and motivated by an incentive to cultivate trust in AI systems \citep{Jacovi}, ML developers have embraced  Explainability (XAI) to develop end-users explanations. 
This work inquires whether, and to what extent, can end-user Explainability satisfy the right to explanation of AI systems' requirements by law. Embracing a "court-like" setting, \secref{sec:legal explanation} makes the case in favor of Explainability, addressing its organic evolvement, its usefulness for ML professionals and its often-cited potential contribution to protecting human rights. Next, \secref{sec:the role} and \secref{sec:xai} lay down the case against end-user Explainability. Accordingly, \secref{sec:the role} sets the legal backdrop by providing a broad-brush analysis of the role of explanations in the legal domain. This analysis focuses on three questions: (1) \emph {How} is "explanation" defined in law, linking it to the notion of "reason-giving"; (2) \emph{ Where} is reason-giving used in law, briefly surveying its appearances in public, private and international law, and most importantly; (3) \emph{ Why} is reason-giving used in law, meaning what are the underlying functions at the heart of the ubiquitous legal practice of reason-giving? 

The analysis of reason-giving's legal functions in \secref{sec:the role} uncovers its four main purposes: (a)~promoting the making of a better and more just decision, (b)~facilitating due-process, (c)~authenticating human agency of both the decision subject and the decision maker, and (d)~enhancing the decision makers' authority, by promoting legitimacy, accountability and providing guidance.
As an interim conclusion, \secref{sec:the role} highlights the fact that reason-giving is a mechanism aimed to influence the human decision maker in various forms, subsequently restraining human rationale and human judgement. 
Building upon this methodology \secref{sec:xai} continues to make the case against Explainability. It utilizes reason-giving's deconstruction in \secref{sec:the role} to analyze the extent to which end-users Explainability is capable of serving the roles assigned to reason-giving in law. It first examines Explainbility's potential to impact the decision-making process itself, finding it slim given that a human decision-maker has been replaced by a prediction-making machine. Thus, reason-giving's function to promote the making of a better and more just decision is largely unfulfilled. Next, the analysis questions the ability of machine generated explanations to support human agency and respect human autonomy. Then, turning to the function of facilitating due-process rights, the analysis highlights Explainability's challenge to produce what is typically considered in law "an explanation". Having ruled out end-user Explainability's ability to serve three of reason-giving's functions in law, we do find that Explainability is compatible with fulfilling reason-giving's fourth function, i.e., enhancing the decision makers' authority. However, we observe that recent Explainable AI (XAI) research trajectories and Large Language Models' (LLMs) emerging capabilities raise serious challenges to the reliability of Explainability's outcomes and create a potential for manipulate end-users by humans and machines alike. 
As a final conclusion, the study outlines some policy implications. The gap between a legal right to explanation and the technological field called Explainability, challenges the usefulness of Explainability as a reason-giving tool in an end-user AI context. Policymakers and ML practitioners should thus reconsider reliance on end-users' Explainability for achieving the societal goals of reason-giving and explore alternatives.

\section{Making the Case for Explainable Artificial Intelligence}\label{sec:legal explanation}
Prior to the legal and regulative interest in a "right to explanation" of AI systems, Explainability was developed by the ML community as a means to contend with one of the most publicly known features of AI systems: it's increasing opacity, or as more commonly known, the "black box" quality. Although not all AI systems are opaque, and there are several "degrees" of opaqueness, this quality nevertheless has become a meaningful challenge due to the introduction of Deep Learning Networks, some of them using billions of parameters. In sync with the rise of the level of complexity of data science, attempts began to offer means to address the opacity challenge, culminating in several concepts, methods, and tools, Explainability (XAI) being one of them \citep{Nicholas}.

The term "Explainability" originates to the 80's and 90's \citep{Miller}. It was developed in order to produce good quality (robust) systems, which consists of an understanding of their inner workings, quality control, bug solving and continuous learning and progress towards the next generation of technology. At its core, XAI "seeks to bring clarity to how specific ML models work" \citep{Laato}, and the use of Explainability is often linked to context and relevancy considerations \citep{Rudin,Molnar,Arrieta}. This fact highlights how the progress made in easing the opacity challenge evolved from real professional challenges: finding out how the system works in order to improve it, fix it, extract takeaways from mistakes and strive to simplify the process \citep{Arrieta}. This core necessity sets the tone for the various technical solutions which were offered and are still being continuously developed to put forward explanations for automated systems, housing a vast amount of research work at the cutting edge of AI today \citep{Biran}. 

Additionally, industry has also acknowledged the problem opacity creates in the general public, asked to be subject to major life-changing and at times high stakes decisions, construed by machines. Clearing out some of the mist around AI systems is often regarded as a step towards creating public trust in this innovative technology \citep{Jacovi}. This approach was largely facilitated by the increased focus of the HCI (Human Computer Interaction) field on extending the definition of human actors interacting with the machine. XAI was embraced by the HCI field at the intersection with the ML community, in a mission to make computational processes clearer to humans \citep{Shneiderman}. Accordingly, explanation in the field of computer science has been understood as "making it possible for a human being (designer, user, affected person, etc.) to understand a result or the whole system" \citep{Malgieri}. This comprehensive definition represents perhaps the turning trajectory of XAI towards including the end user of AI systems, re-calibrated in correlation with the increasing deployment of these systems in domains already regulated by existing laws. But it also highlights the fact that explanations for AI systems are often mentioned in the context of the mission to promote trust, or trustworthiness, in AI \citep{Laato}. The absence of an ability to explain decisions and actions by AI black-boxes to human users has been recently referred to as a key limitation of today's intelligent systems, whereby the "...lack of Explainability hampers our capacity to fully trust AI systems"\citep{Mehta}. And trust, it has been argued, promotes user's utilization of models, both by relying on its predictions as well as by accepting its deployment \citep{Ribeiro}.

Against this technological backdrop, Explainability has been enlisted to secure a legal mechanism, the so called "right to explanation", which regulators sought for the protection of society from potential AI harms.  Regulation of AI systems grasped the vast potential of automated systems on the one hand, but expressed a genuine concern towards safeguarding human rights  \citep{OECD}. Being preoccupied with the purported "black-box" quality of AI systems, regulation sought transparency enhancing mechanisms to address those concerns. In the legal domain, transparency is often linked to fairness, as means to assure accountability of decision makers \citep{Kaminski2021B}. Faithful to this transparency ethos, "...the majority of discourse around understanding machine learning models has seen the proper task as opening the black box and explaining what is inside" \citep{Selbst}. Accordingly, explanations for AI systems are being promoted in service of multiple regulatory objectives aiming at enhancing transparency. Thus, explanation-giving for AI systems was mentioned as means for achieving AI accountability\citep{Doshi-Velez,Smith-Renner,Gillis}, detect discrimination \citep{Brkan}, reveal bias issues \citep{Melsion}, and ensure fairness in AI systems \citep{Dodge}. In regards to governmental use of AI, explanation-giving is presented as a way to accommodate due process requirements and achieve good governance \citep{Crawford}. Similarly, it is also considered essential in order to allow for a meaningful contestation right towards automated decisions \citep{Kaminski2021A}.

This extensive list highlights the diverse groups, interests, and contexts for which a right to explanation of AI systems is considered a desired feature, as well as demonstrates the large extent of reliance on transparency in general, and explanations in particular, by regulators, legal practitioners, and legal scholars. This insight prompts the following question: why did regulators and legal practitioners turn to the tool of explanation-giving, in service of protecting humanity against AI harms? The answer lies in the role of explanations in law and law's ubiquitous use of explanations.
\section{The Role of Explanation in Law - "What" is Explanation in Law?}\label{sec:the role}
"The business of law is the business of making decisions"\citep{Hawkins}. This eloquent statement captures the fact that decision-making resides at the heart of the legal system. In a democratic society decision-making is often accompanied by explanations of those decisions \citep{Rawls}, making it a common practice both for law-making and law-applying \citep{Raz}. This form of "reason-explanations", typically used when humans try to understand and explain action and resolve disagreements \citep{Baum}, is usually referred to as "reason-giving". Its use is so ubiquitous that "the practice of providing reasons for decisions has long been considered an essential aspect of legal culture" \citep{Schauer}. To deconstruct the notion of reason-giving in law and to answer the question "\emph{ What} is reason-giving in law?", this section will ask the following questions: \emph{ How} is reason-giving defined in law? \emph{Where} can we find the use of reason-giving in law? and most importantly \emph{Why} does law uses reason-giving to begin with, meaning what are its underlying functions?
\subsection{Reason-Giving in Law - the "How" - Defining Key Terms} 
In order to alleviate some of the "fuzziness" around basic concepts it is important to first define their meaning. The giving of reasons can be described as "the practice of engaging in the linguistic act of providing a reason to justify what we do or what we decide" \citep{Schauer}. The difference between explaining ("providing a reason") and justifying ("to justify") is not strictly semantic. While explanation in a general sense means "an act of spotting the main reasons or factors that led to a particular consequence, situation, or decision" \citep{Malgieri}, a justification takes on another layer, detailing why the decision at hand is the "right" and "just" one \citep{Malgieri}. Therefore, an explanation is part of the justification. Explanations and justifications will be collectively referred to here as reason-giving, the process whereby decisionmakers elaborate the explanations and justifications supporting their decisions \citep{Deeks}. Indeed, reason-giving is particularly suitable for legal decisions since "[w]hen we provide a reason for a particular decision, we typically provide a rule, principle, standard, norm, or maxim broader than the decision itself..." \citep{Schauer}. It should be noted here that reason-giving has a multi-layered presence in law. For example, law demands reason-giving (e.g., courts requiring agencies to produce reasons for a decision), and law also manufactures reasons simultaneously (e.g., courts justifying their rulings re the agencies' actions). Moreover, reason-giving is relevant both as part of the decision-making process itself (adjudicating - the process of deliberating and deciding), and as a product accompanying the final decision if released publicly.
\subsection{Reason-Giving in Law - the "Where"}
Reason-giving and explanations are being ubiquitously used across the legal system. Some of the most dominant arenas where the legal system leverages reason-giving are public law, private law, and increasingly international law. In a nutshell, public law is perhaps the most widely recognized domain of reason-giving in the legal system, construed out of courts, agencies and legislators constantly manufacturing and reviewing explanations and justifications. Private law exemplifies the extent through which "regulatory transparency" has become the tool-of-choice to handle regulatory challenges \citep{Weil}, where perhaps the most prominent example is the requirement to obtain a patient's informed consent prior to undergoing medical procedures, itself contingent upon receiving an explanation from a physician \citep{McLean}. Civil law also entails numerous examples of explanation usages such as in contractual relationships or in Tort law. In addition, the newly emerging habit of nations to explain foreign policy as part of international law fortifies the importance of reason-giving to decision making as a legal and social phenomenon, transcending states and geolocations\citep{Keitner, Kingsbury}. 
\subsection{Reason-Giving in Law - the "Why"}
The answer to the question "what does society gain from this constant explanation giving?" ought to spearhead the methodological framework of end user Explainability. Accordingly, this section will detail why are explanations and reason-giving such a repetitive practice in law? What purposes do they serve and what are their underlying functions?
\begin{enumerate} 
\item {\bf Making a Better and More Just Decision} - At the heart of reason-giving in law lies the non-instrumental purpose of securing a better and more just decision \citep{Deeks}. The "just" feature, which supports the act as right, desirable, or reasonable, authenticates the decision as a non-biased, non-discriminatory one \citep{Gillis}. It taps into the core objective of making sure that "justice was done" \citep{Atkinson}. The "better" feature is brought to fruition by triggering the mechanism of review, either internal during the making of the decision, or external as means for appeal and contestation. Taken together, the decision possesses both a rational and a moral basis, making it a more righteous and fair result. In this sense, there's an inherent, non-instrumental value in reason-giving since it impacts the decision itself. Reason-giving compels the decision maker to handle the decision process with extra care, in a thoughtful and slower manner. There might also be a psychological pressure on decision makers to make decisions worthy of reasonable reasoning \citep{Cohen38, Shapiro}. In other words, the need to articulate reasonable reasoning for a decision nudges decision-makers to make decisions that support such reasoned reasoning in a circular movement. Therefore, the mere fact that reasoning may be required may impact the decisions even prior to such a request being materilized. 
\item {\bf Facilitating Due Process} - understanding the decision-making system has been said to be instrumental for individuals to exercise their right to challenge decisions \citep{Gillis}. When focusing on its role as a protector of individual rights, a right to explanation is usually regarded as a parasitic right, in service of fulfilling other values \citep{Cohen40, Mashaw2007}. Those include a right for due process, housing both a right to a hearing \citep{Friendly} and a right to contestation \citep{Kaminski2021A}. The due-process theory is a core principle of the rule of law itself \citep{Kaminski2021A}, and the procedure of due process is referred to today mainly as the requirement that any infringement on core rights should be taken after a notice was given and an opportunity for a hearing was granted \citep{Crawford}. Reason-giving plays multiple roles in the execution of due process rights. Naturally, knowing one's reasons for a decision assists in crafting better informed arguments to rebuttal it \citep{Cohen38}, thus supporting a robust defense against a rights-infringing decision or act. Of course, due process allows for a judicial review of the decision and is especially instrumental given its contribution to the conservation of records which can be leveraged later for contestation and review of the aforementioned decision or action. Moreover, it allows the decision maker and contesting party to evaluate the chances of an appeal in advance. Finally, the giving of reasons can serve as a non-political legitimate demand by the adjudicating body, in comparison to a more subjective requirement of decision "reasonableness".
\item {\bf Acknowledging Human Agency of the Decision Subject and Decision Maker} - One of the core values underlying the existence of reasons for decisions is respecting human autonomy \citep{Gillis}. In the case of the \emph{ decision-subject}, the reasons issued to a decision signal his or her sovereignty, since giving reasons respects the fact that humans are autonomous people that should be treated with dignity, while unreasoned coercion "denies our moral agency and our political standing" \citep{Mashaw2007}. Moreover, respect comes in the form of providing grounds for detailed criticism not only when there is a right for contestation, but perhaps even more when there is no recourse for appeal (e.g., reason-giving accompanying Supreme Court decisions) and a decision subject is left with a right to public discourse. Additionally, reasons respect also the \emph{ decision-maker's} human agency. At that capacity of a human decision maker, the presence of reasons for one's actions and decisions stands at the heart of human morality and sense of judgment and autonomy \citep{Mashaw2007}. Plainly put, a human decision maker needs there to be reasons for its actions, as an autonomous person. When actions are underlined with intent, the decision maker is acting as a rationale agent, thus strengthening his or her autonomy in the process. 
\item {\bf Enhancing the Decision Makers' Authority} - The giving of reasons makes actions, decisions, rules, and regulations more tolerable and acceptable. This is because acknowledging them as binding is dependent upon there being sufficient rational explanations underlying those rules \citep{Mashaw2001}. Simply put, "the authority of all law relies on a set of complex reasons for believing that it should be authoritative" \citep{Mashaw2001}. Reason-giving contributes to this objective by supporting attributes that promote compliance and adherence to the deciding body. These attributes comprise of enhancing the accountability and legitimacy of the deciding body, as well as the providence of guidance to numerous stakeholders (while simultaneously serving as a binding precedent on the decision maker itself). Those virtues jointly add to maintaining and boosting agreement, cooperation and acceptance of rules established by the decision-making body, thus bolstering the system's mandate. They also serve as a pressure system of socio-legal and relational considerations cast upon the human decision maker, which is often concerned with matters of reputation, colleagues' approval, avoidance of unpleasant repercussions when reviewed, and various other incentives to make the "right" decision and provide meaningful explanations to reason it \citep{Mashaw2007}.  
\end{enumerate}
From executing a right for due process, contributing to the making of a better and more just decision, respecting human agency and promoting the decision makers' authority, reason-giving's central role in law serves purposes oriented towards the decision subject, but also to a larger degree towards the human decision maker. Leveraging the existence of societal and relational pressures upon the human decision maker, reason-giving is a legal tool aimed primarily to contain, retrain, and curb human discretion and human judgement. This conclusion is also supported by instances where a requirement for explanations in law is absent, as is the case of jurors \citep{Doshi-Velez}, where the value of restraining human judgement is attained by other means, such as internal deliberations. 
Having outlined reason-giving's role in law, being a basis for the regulatory pursuit of a right to explanation towards AI systems, it is now possible to ask to what extent can Explainability execute reason-giving's role in law and society? In other words, can Explainability successfully fulfil reason-giving's functions?  
\section{Can End-User Explainability Fulfil a "Right to Explanation"?}\label{sec:xai}
"Explainability" has really come to dominate debates about ethics and regulation of machine learning \citep{Bordt}, largely framed as the tool to execute a right to explanation of AI systems. As several survey papers demonstrate \citep{Adadi,Carvalho,Guidotti}, there is considerable effort employed at identifying a suitable framework or methodology for XAI in the context of end-users \citep{Arrieta,Langer,Prakken,Tomsett}. However, currently, and despite this formidable effort, scholars have pointed out that the tool of Explainability is mostly used for professional debugging purposes \citep{Mittelstadt}, and has not yet managed to translate into a user-friendly explanation generating tool, albeit regulatory calls for an individual, decision-subject right to explanation \citep{Goodman2022}. Since "...much work in AI and ML communities tends to suffer from a lack of usability, practical interpretability and efficacy on real users" \citep{Ashraf}, Explainability for end-users is proving to be a tough challenge. As scholars recently lamented, "...so far at least, aspirational Explainability cannot be relied upon either for effective communication about how algorithmic systems works or for holding them to account" \citep{Goodman2022}.

Leveraging the legal reason-giving methodology presented in \secref{sec:the role} , this section proposes to frame the persistent gap between a right to explanation and Explainability, by examining to what extent can end-user Explainability fulfil the role law bestows upon explanations and reason-giving, and will accordingly ask: (a)~can it contribute to the making of a better and more just decision? (b)~can it facilitate due-process rights? (c)~is it relevant for the authentication and respect of human agency, and (d)~does it enhance the decision makers' authority? 
\subsection {Can Explainability Contribute to a Better and More Just Decision?}
If one of reason-giving's main roles is to impact the decision-making process itself by restraining human judgement and thus contribute to a better and fairer decision, then it is hard to grasp in what form this purpose might be fulfilled given a machine now replaces a human decision maker. The impact of reason-giving on humans, slowing down decision processes and leveraging relational pressures, is largely irrelevant when a machine's decision is involved. Unlike a human decision maker, reason-giving does not serve to contain an algorithm's judgement or discretion. An algorithm does not possess a "rational" (or logic) to begin with, nor does it produce a "decision" but rather a prediction. It is not impacted, nor impressed, by what other algorithmic colleagues may think of it, nor does it seek to minimize unpleasant consequences, or "feel" accountable to anyone or anything. Therefore, currently prediction algorithms make no use of the external explanation generated to their predictions. There might be some potential impact over humans in the "surroundings" of a model, e.g., designers, deployers etc., but this impact, if exists, should be further explored, and is probably diminished. Therefore, it appears that one of the most important objectives of reason-giving cannot be attained using Explainability for end-users. 

\subsection{Is Explainability Instrumental to Facilitating Due-Process Rights?}
To facilitate reason-giving's decision-subject purposes such as due process rights, appeal and contestation, this work proposes that Explainability should deliver decision-subjects with what law considers to be "an explanation", and reliable at that. 
\subsubsection{Can Explainability Generate "an Explanation"?}
In a call to stay clear of black-box models, one of the more significant scholars in the field of ML, \citep[Rudin], has opined that "[a]s the term is presently used in its most common form, an explanation is a separate model that is supposed to replicate most of the behavior of a black box..." . In essence, the general concept dominating the XAI community is "to create a simple human-understandable approximation of a decision[...]making algorithm that accurately models the decision given the current inputs..." \citep{Wachter}. These insights frame the different mehtods that were developed over the years to provide explanations for models, such as LIME, SHAP, LRP, etc. \citep{Linardatos} and hints at the inadequacy of calling their output "an explanation" in nomenclature, as suggesting a reliable knowledge of how the complex model works \citep{Mittelstadt}. 

\begin{figure}[tb]
\includegraphics[width=0.9\textwidth]{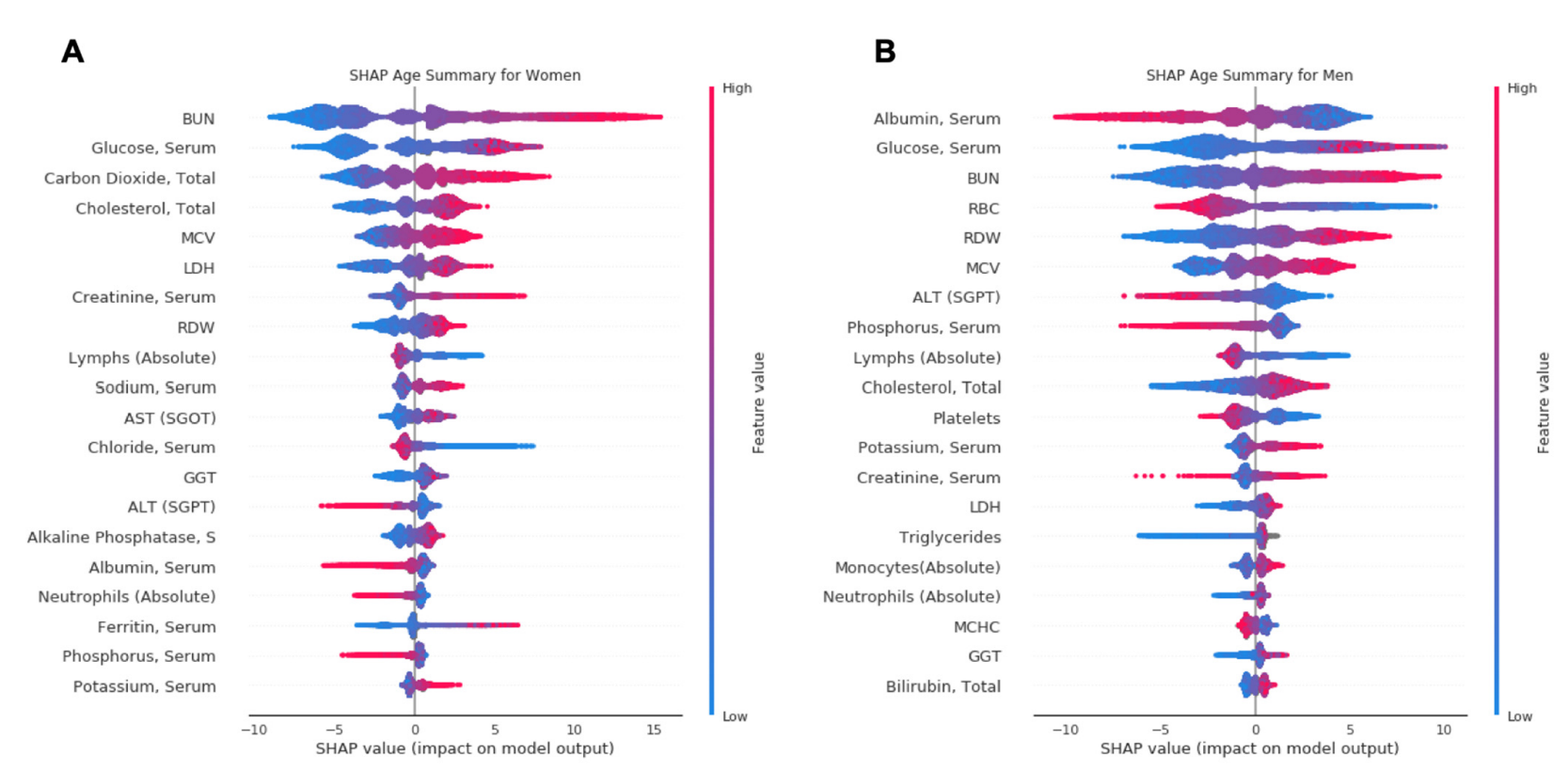}

\caption{An example of SHAP Explainability technique summary plots from \citet{ wood2019interpretable}\label{fig:shap}}
\end{figure}

In fact, those "explanation-generating" techniques should be regarded as producing a clue to the source of the issue explored, by providing vague approximations of how the algorithm generated its output or some understanding of the features that need to be changed in order to alter the said output \citep{Bordt}. This insight requires further inquiry and human deduction skills, given causality may not be automatically inferred from the data an explanation has provided. It is up to ML experts to then leverage this clue and find the true cause for the decision/problem itself  \citep{Mittelstadt}. 
True for ML experts, this is double the case for a layperson lacking technological background. Even if Explainability techniques can produce an actual contextualized explanation rather than a clue, scholars argue it is still a long way from producing layperson understandable explanations \citep{Umang}. In fact, most current Explainability techniques are non-accessible to a human lacking technological literacy \citep{Wachter}. As \figref{fig:shap} demonstrates A\emph{ run-of-the-mill} person would have slim understanding of a saliency map, a data points analysis, or a feature importance result. Some may struggle even to understand a bar chart. Therefore, some kind of brokerage work would be needed, where a trusted expert would have to translate Explainability technique results to a person seeking an actual meaningful explanation. In this case, users' trust will be built upon experts' opinions rather than end user explanations, similar to many experiences in our lives like trusting the functioning of a navigation compass or trusting an engineer while crossing a bridge, where trust is granted not based on an explanation but on other features \citep{Jarek}.

Based on this examination it appears Explainability is currently not sufficient to deliver what regulators consider "an explanation". But even if it could deliver on such a requirement, can Explainability be trusted by decision-subjects to begin with?
\subsubsection{Can Decision Subjects Rely on Explainability Generated "Explanations"?}
If users and decision subjects cannot rely on the explanations generated by end-user Explainability, then a major obstacle hinders its adoption. Research in the field has highlighted few potential problems in this regard. First, not all stakeholders tasked with generating explanations to automated decisions welcome this explanation generating requirement. Some concerns can include potential infringement of privacy rights, intellectual property and trade secrets, and genuine security concerns \citep{Wachter,Powell,Smitha,Tramer}. Additionally, a potential to game the system when receiving explanations on the one hand, or a perceived inability of end-users to comprehend complex systems on the other hand, might also contribute to designers' resentment towards end-user Explainability \citep{Powell,Zhang}. And sometimes models are just so complex that it is claimed they simply cannot be explained in a meaningful way \citep{Jarek}. 
One should also not overlook the inherently adversarial relationship between end-users and automated decision-maker stakeholders, given that end-users and automated decision subjects largely seek an explanation to change the machine's prediction (e.g., loan-seeker vs. credit score generator). Adversarial situations invite ambiguous and non-trustworthy explanations to begin with \citep{Botty}, and there are multiple techniques to possibly manipulate the "explanation" generated by Explainability methods \citep{Bordt,Joyce,Mothilal}. 
\subsection{Can Explainability Authenticate Human Autonomy?}
Naturally, the change of the decision makers' identity, meaning an autonomous decision-making system vs. a human decision maker nulls reason-giving's function as an acknowledger of the decision makers' human agency. However, we believe that end-user Explainability's potential to acknowledge the decision subject's humanity and autonomy should also be questioned. While residing outside the scope of this work, this function raises multiple important and fundamental moral and philosophical questions relevant to AI systems in general, and to XAI in particular (e.g., can human agency be acknowledged by a non-human agent to begin with?). 

So far, we have examined end-user Explainability's ability to fulfil three of reason-giving's functions in law and found it lacking. Turning to the final function, we at last find a function which Explainability is well suited to deliver, a fact which simultaneously raises serious concerns.  
\subsection{Can Explainability Enhance the Decision-Makers' Authority?}
\begin{figure}[tb]
\includegraphics[width=0.9\textwidth]{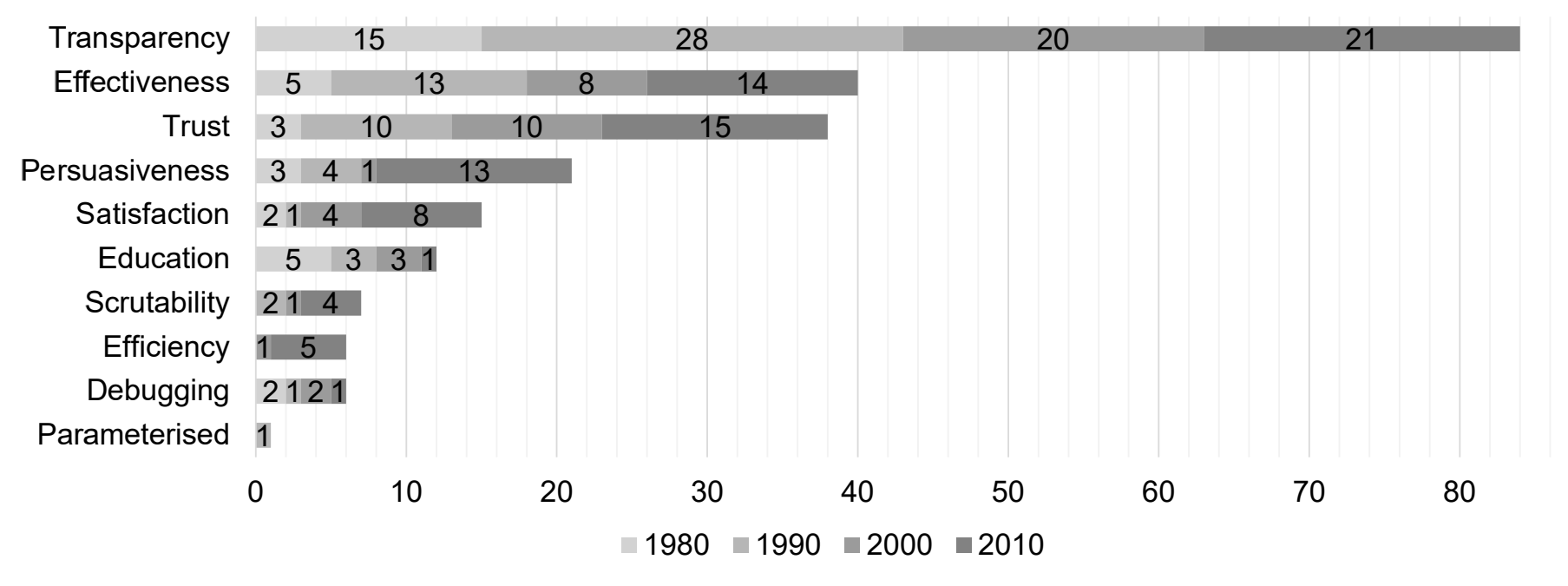}

\caption{\citep{Nunes} The figure shows that past decades demonstrate a plateau, or even a decrease, in researching explainability techniques for the purpose of transparency ("explain how the system works") and a sharp increase in purposes such as enhance explanations' effectiveness ("help users make good decisions"), enhance trust ("increase users' confidence in the system") and enhance persuasiveness ("convince users to try or buy"). Although this paper's survey dates to 2017, it is plausible to assume that a current overview will demonstrate an even stronger orientation towards user influencing purposes. All purposes definitions are taken from Table 8 of the surveyed paper. \label{fig:nunes}}
\end{figure}

Contributing to the decision-makers' authority and legitimacy is another function of reason-giving in law. In the case of end-user Explainability, we find this function can be successfully fulfilled, perhaps even better than human decision makers, especially in the case of LLMs. As a recent paper exploring GPT-4's explanation's abilities demonstrates, it "is remarkably good at generating reasonable and coherent explanations, even when the output is nonsensical or wrong" \citep{Bubeck}.
However, we recognize several problems emerging from XAI's ability to promote the decision makers' authority. First, at the heart of this function lies reason-giving's impact on the human decision maker. This impact means he or she feels accountable, seeks legitimacy and is bound by his or her previous decisions if they are to serve as guidance. Therefore, the replacement of a human with a machine nullifies most, if not all, of the aforementioned human effects.
Moreover, the "explanation" Explainability generates is limited in the sense that we inherently expect an explanation to be based on some knowledge of the world (contextualized), whereas an algorithm only "knows" (if one can even attribute such adjective to a machine) what it was shown or defined to "know" \citep{ Bordt, Lipton}. In other words, and until Artificial General Intelligence proves otherwise, "[e]very AI system is the fabled tabula rasa; it "knows" only as much as it has been told" \citep{Jarek}. Under these conditions, an explanation cannot function as a rule, nor as guidance.
Equally disconcertingly perhaps, although a human decision maker is increasingly replaced by a machine, one fact has yet to change, and that is the human identity of the decision subject. In automated decision making, a human is still the client/target of the explanation, a matter which potentially gives rise to rather alarming consequences. Research has shown Explainability's potential to cause human over-reliance on the system \citep{Smith-Renner}, as well as the opportunity for wrongdoing and manipulation by promoting misguided trust. This phenomenon of nudging users to act according to others' interest is known as "Dark Patterns" in XAI \citep{Gray}, and benefits from humans' "automation bias" towards trusting machines \citep{Malin,Kaminski2021A,Lyell}. For example, people are more eager to comply with a request simply by being presented a placebic justification by computerized systems \citep{Malin}. Further research has suggested that user manipulation can occur even unintentionally, causing "Explainability pitfalls" merely by choosing to present people with one sort of explanation over another \citep{ehsan}.
It should also be pointed out that end users XAI appears to drift away from its initial trust building objective. For example, \citep{Laato} have shown that transparency is mostly evaluated in the literature according to the user's perception of transparency, rather than actual transparency attributes of the system. As a systematic review of papers in the field conveys, research in the field scarcely highlights the purpose for generating explanations to begin with \citep{Nunes}. Moreover, it appears the research of end-users' XAI is increasingly shifting towards exploring which explanation practices will impact users' trust and increase perceived trustworthiness of the system, rather than produce a meaningful and reliable tool to scrutinize AI systems by users \citep{Forster}. As \figref{fig:nunes} taken from \citep{Nunes} demonstrates, surveying hundreds of XAI papers in the last decades displays a plateau or even an overall decrease in the study of XAI for transparency purposes, and a big increase in researching explanations' effectiveness, explanation techniques to enhance user's trust, techniques to increase explanations' persuasiveness, and to elevate user's levels of satisfaction from the system.  
Two interesting examples of this trend include \citet{Weitz} demonstrating how the use of virtual agents for Explainability seems especially promising for the purpose of increasing users' perceived trust in the system, or \citet{Goldman} experimenting with explanations as a technique to elevate user's comfort level in automated driving maneuvers of a simulated autonomous vehicle, to avoid manual take-overs. Both examples showcase how end-users' XAI might drift away from its original purpose of using explanations to promote "appropriate trust" \citep{Gunning} and assist end-users to properly scrutinize AI systems \citep{Forster}, towards the study of how XAI can be used to influence end-users according to third parties' incentives, well intended as they may be. 
Finally, the emerging capabilities of LLMs demonstrate that machines might pose a risk when pursuing end-user Explainability. Recent models such as GPT-4 exhibit an increasing ability to generate convincing explanations for false decisions, a lack of consistent link between the decision-making process and the explanation generated to it, and a growing capability to generate specially tailored explanations to a human client \citep{Bubeck}. As Turpin et al recently demonstrated, LLMs can produce step-by-step reasoning which systematically misrepresents the real reason underlying the model's prediction \citep{Turpin}. Therefore, there is real potential to contribute to the decision-making system's trustworthiness, even when that trust is an unwarranted, misleading and even a dangerous one. These continuously improving capabilities should serve as a trigger warning for those promoting end user explanations.

\section{Conclusion}\label{sec:conclusion}
The deconstruction of reason-giving in the legal system this study presents offers a methodological framework to analyze the gap between a right to explanation and end-user Explainability. It highlights reason-giving's role as impacting the human decision maker, as well as facilitating decision subjects' rights. Given the change in the identity of the decision maker from human to machine, current end-user Explainability struggles to deliver most of explanations' functions in law, which include promoting a better and more just decision, facilitating due-process and acknowledging human agency. In contrast, end-user Explainability emerges as a successful mechanism to fulfill reason giving's fourth function in law, i.e., enhancing the decision makers' authority. However, this ability raises a set of risks for manipulating decision subjects by humans and machine alike. A key limitation of the case against Explainability is that it does not yet provide an alternative solution to the risks stemming from recent AI advancements \citep{Bubeck}. Nevertheless, we fear the reliance on inadequate techniques, coupled with newly generated risks, is perilous. Hence, we hope our work will impact how Explainability is being developed and implemented and will serve as a warning sign from incompatible usage and unwarranted research directions. 
\section*{Acknowledgement}
This work has been supported by the Israeli Science Foundation research grant 1437/22 and a
grant from the Tel Aviv University Center for AI and Data Science (TAD).

\bibliography{XAI}

\end{document}